\documentclass[final]{IEEEtran}

\usepackage{times}
\usepackage{epsfig}
\usepackage{graphicx}
\usepackage{amsmath}
\usepackage{amssymb}
\usepackage{enumitem}
\usepackage{subfigure}
\usepackage{booktabs}
\usepackage{multirow}
\usepackage{bm}

\usepackage[pagebackref=false,breaklinks=true,colorlinks,bookmarks=false,hidelinks]{hyperref}

\begin{document}

\title{Low-Shot Learning for the Semantic Segmentation of Remote Sensing Imagery} 

\author{Ronald~Kemker,~Ryan~Luu,~and~Christopher~Kanan,~\IEEEmembership{Member,~IEEE}
\thanks{R. Kemker, R. Luu, and C. Kanan are with the Machine and Neuromorphic Perception Laboratory in the Carlson Center for Imaging Science, Rochester Institute of Technology, Rochester, NY, 14623 USA (http://klab.cis.rit.edu/).}


}


\maketitle

\begin{abstract}
Recent advances in computer vision using deep learning with RGB imagery (e.g., object recognition and detection) have been made possible thanks to the development of large annotated RGB image datasets.  In contrast, multispectral image (MSI) and hyperspectral image (HSI) datasets contain far fewer labeled images, in part due to the wide variety of  sensors used. These annotations are especially limited for semantic segmentation, or pixel-wise classification, of remote sensing imagery because it is labor intensive to generate image annotations. Low-shot learning algorithms can make effective inferences despite smaller amounts of annotated data. In this paper, we study low-shot learning using self-taught feature learning for semantic segmentation.  We introduce 1) an improved self-taught feature learning framework for HSI and MSI data and 2) a semi-supervised classification algorithm. When these are combined, they achieve state-of-the-art performance on remote sensing datasets that have little annotated training data available.  These low-shot learning frameworks will reduce the manual image annotation burden and improve semantic segmentation performance for remote sensing imagery. 
\end{abstract}


\IEEEpeerreviewmaketitle

\section{Introduction}

\IEEEPARstart{S}{emantic} segmentation is a computer vision task that involves assigning a categorical label to each pixel in an image (i.e., pixel-wise classification). For color (RGB) imagery, deep convolutional neural networks (DCNNs) are continually pushing the state-of-the-art for this task. This is enabled by the availability of large annotated RGB datasets. When small amounts of data are used, conventional DCNNs generalize poorly, especially deeper models. This has made it difficult to use models designed for RGB data with multispectral imagery (MSI) and hyperspectral imagery (HSI) that are widely used in remote sensing, since publicly available annotated data is scarce. Due to the limited availability of annotated data for these ``non-RGB'' sensors, adapting DCNNs to remote sensing problems requires using low-shot learning. Low-shot learning methods seek to accurately make inferences using a small quantity of annotated data. These methods typically build meaningful feature representations using unsupervised or semi-supervised learning to cope with the reduced amount of labeled data. 

Many researchers have explored unsupervised feature extraction as a way to boost performance in semantic segmentation of MSI and HSI.  They have tried  shallow features (e.g., gray-level co-occurrence matrices~\cite{mirzapour2015improving}, Gabor~\cite{shen2013discriminative}, sparse coding~\cite{soltani2015pixels}, and extended morphological attribute profiles~\cite{ghamisi2014automatic}), and deep-learning models (e.g., autoencoders~\cite{liu2015hyperspectral, lin2013spectral, zhao2015combining, ma2015hyperspectral, tao2015unsupervised}) that learn spatial-spectral feature extractors directly from the data.  Recently, self-taught learning models have been introduced to build feature-extracting frameworks that generalize well across multiple datasets~\cite{kemker2017self}.  In self-taught learning,  spatial-spectral feature extractors are trained using a large quantity of unlabeled HSI and then used to extract features from other datasets that we may want to classify (i.e. the target datasets). Self-taught learning for HSI semantic segmentation was pioneered in \cite{kemker2017self}. 

\begin{figure}[t]
    \centering
    \includegraphics[width=0.99\linewidth]{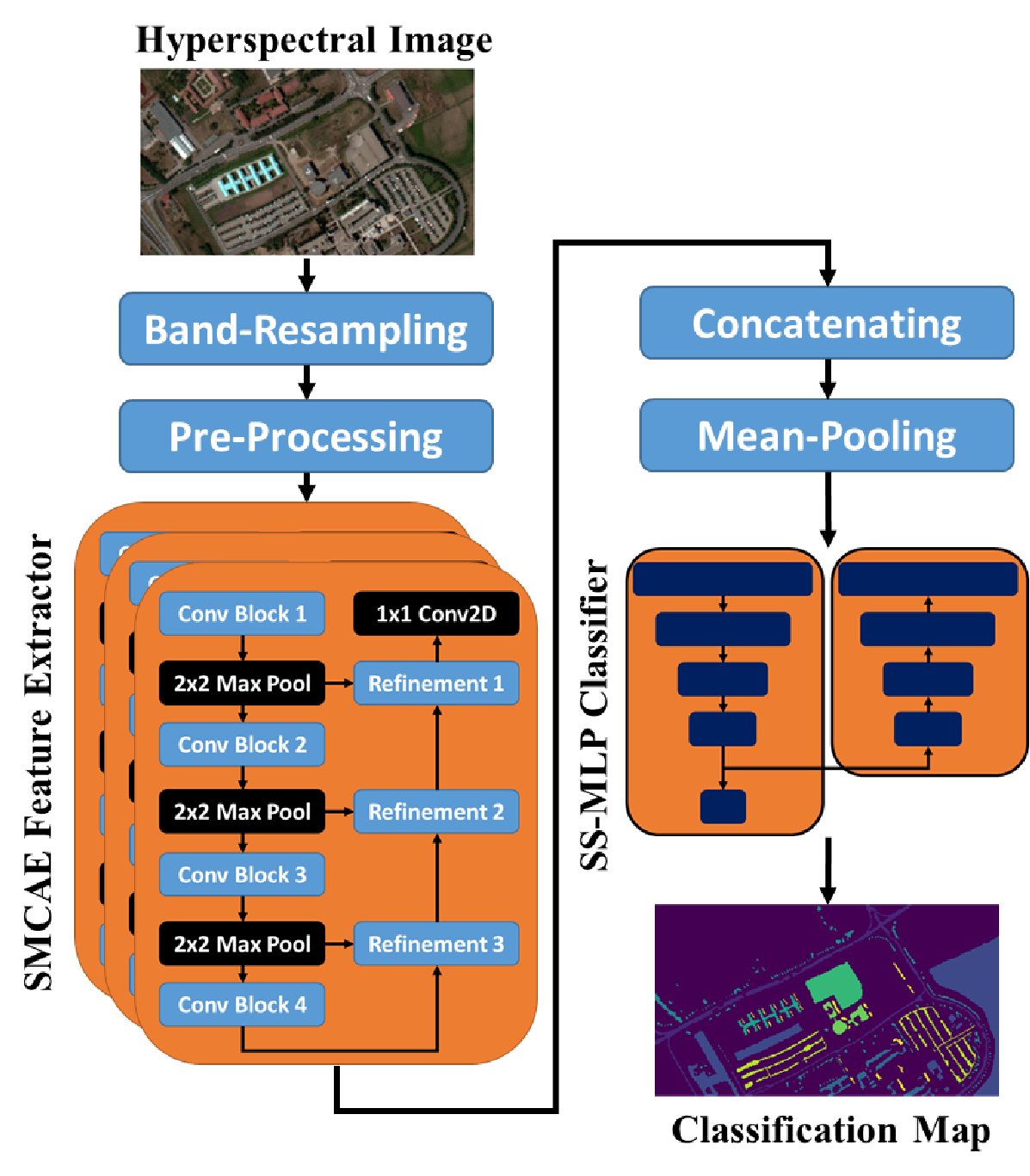}
    \caption{Our proposed SuSA architecture for semantic segmentation of remote sensing imagery. For feature extraction, SuSA uses our SMCAE model, a stacked multi-loss convolutional autoencoder  that has been trained on unlabeled data using unsupervised learning. For classification, SuSA uses our semi-supervised multi-layer perceptron (SS-MLP) model.\label{fig:upfront}}
\end{figure}

As the dimensionality of each feature vector increases, the performance for many deterministic models (e.g., support vector machine (SVM)) will degrade~\cite{bellman2013dynamic}.  The most common method for preventing this is to reduce the dimensionality of the feature space (e.g., using principal component analysis (PCA)); however, this involves tuning at least one more hyperparameter (i.e., number of dimensions to retain) through cross-validation.

Multi-layer perceptron (MLP) neural networks can learn which features are the most important for classification; however, they normally require a large quantity of annotated data to generalize well.  Semi-supervised learning uses an unsupervised task to regularize classifiers that do not have enough annotated data to work with.  For example, the ladder network architecture proposed by Rasmus et al.~\cite{rasmus2015semi} trains on labeled and unlabeled data simultaneously to boost segmentation performance on smaller training sets.  Semi-supervised frameworks give the model the ability to increase the dimensionality in the feature space, which allows them to learn what features are most important for optimal performance, and also enables them to perform well with little annotated data.

In this paper, we describe the semantic segmentation framework SuSA (\textbf{s}elf-ta\textbf{u}ght \textbf{s}emi-supervised \textbf{a}utoencoder) shown in Fig.~\ref{fig:upfront}. SuSA is designed to perform well on MSI and HSI data where image annotations are scarce. SuSA is made of two modules. The first module is responsible for extracting spatial-spectral features, and the second module classifies these features.

We evaluated SuSA across multiple training/testing paradigms, and we compared our performance against state-of-the-art solutions for each respective paradigm found in literature, including two recent self-taught feature learning frameworks: MICA-SVM and SCAE-SVM~\cite{kemker2017self}.  We describe these in more detail in later sections.

\textbf{This paper's major contributions are}:
\begin{itemize}
\item We describe the stacked multi-loss convolutional auto-encoder (SMCAE) model (Fig.~\ref{fig:smcae_model}) for spatial-spectral feature extraction in non-RGB remote sensing imagery. SMCAE uses unsupervised self-taught learning to acquire a deep bank of feature extractors. SMCAE is used by SuSA for feature extraction.  
\item We propose the semi-supervised multi-layer perceptron (SS-MLP) model (Fig.~\ref{fig:ssmlp}) for the semantic segmentation of non-RGB remote sensing imagery. SuSA uses SS-MLP to classify the feature representations from SMCAE, and SS-MLP's semi-supervised mechanism enables it to perform well at low-shot learning.
\item We demonstrate that SuSA achieves state-of-the-art results on the Indian Pines and Pavia University datasets hosted on the IEEE GRSS Data and Algorithm Standard Evaluation website.
\end{itemize}

\section{Related Work}
\label{section:related_work}

\subsection{Self-Taught Feature Learning}

The self-taught feature learning paradigm was recently introduced as an unsupervised method for improving the performance for the semantic segmentation of HSI~\cite{kemker2017self}.  In the past, researchers learned spatial-spectral features directly from the target data and then passed them to a classifier~\cite{ma2015hyperspectral,mirzapour2015improving,tang2015hyperspectral,tao2015unsupervised}.  Learning spatial-spectral features on a per-image basis is computationally expensive, which may not be ideal for near-real-time analysis.  Self-taught feature learning uses large quantities of unlabeled image data to build discriminative feature extractors that generalize well across many datasets, so there is no need to re-train these types of feature extracting frameworks~\cite{raina2007self}.

The authors in \cite{kemker2017self} introduced two self-taught learning frameworks for the semantic segmentation of HSI.  The first model, multi-scale independent component analysis (MICA) learned low-level feature extracting filters corresponding to bar/edge detectors, color opponency, image gradients, etc.  The second model, the stacked convolutional autoencoder (SCAE), is a deep learning approach that is able to extract deep spatial-spectral features from HSI.  These pre-trained models would extract features from the source image (i.e., the image we want to classify) and pass them to a support vector machine (SVM) classifier.  Since MICA-SVM and SCAE-SVM provide state-of-the-art performance across multiple benchmark datasets, we compare our proposed work against them. 

The SCAE model consisted of three separate convolutional autoencoder (CAE) modules trained in sequence.  The training loss for each CAE was the mean-squared error (MSE) between the input data and the reconstructed output (also known as the data layer).  It was shown in \cite{valpola2015neural} that backpropagation is better at optimizing trainable parameters that are closer to where the training loss is computed (i.e., training error signal) than the trainable parameters in deeper layers.  The solution was to take a weighted sum of the reconstruction loss for every encoder/decoder pair, which allowed the network to reduce reconstruction errors that occur in deeper layers.  

In this paper, we introduce the stacked multi-loss convolutional autoencoder (SMCAE) spatial-spectral feature extracting framework (Fig.~\ref{fig:smcae_model}).  It is made up of multiple MCAE modules, where each uses multiple loss functions to incorporate and correct reconstruction errors from both shallow and deeper CAE layers.  SMCAE trains, extracts, and concatenates feature responses from the individual MCAEs in the way SCAE is built from individual CAEs.  SMCAE allows the user to extract deep spatial-spectral features directly from the image data.

\subsection{Semi-Supervised Learning}

Self-taught feature learning focuses on unsupervised learning of features on additional data, and then use these features with a supervised system. Semi-supervised algorithms use supervised and unsupervised learning to improve generalization on supervised tasks; which in turn, improves classification performance on test data~\cite{valpola2015neural, salimans2016improved, rasmus2015semi}. In both cases, unsupervised learning helps these algorithms to avoid overfitting when given only a small number of labeled HSI samples.

A number of discriminative semi-supervised methods have been adapted for HSI classification. The transductive support vector machine (TSVM) is a low-density separation algorithm that saw early success. TSVM seeks to choose a decision boundary that maximizes the margin between classes  using both labeled and unlabeled data~\cite{bruzzone2006novel}. TSVM outperformed the inductive SVM when evaluated on the Indian Pines HSI dataset~\cite{bruzzone2005transductive}. TSVM is computationally expensive and has a tendency to fall into a local minima.

Camps-Valls et al.~\cite{camps2007semi} trained graph-based models for HSI classification using labeled and unlabeled data.  Their model iteratively assigned labels to unlabeled pixels that were clustered near labeled pixels. Their model outperformed a standard SVM on the Indian Pines dataset.  Using manifold regularization, the Laplacian support vector machine (LapSVM) expanded the graph-based model and showed promise in MSI classification and cloud screening~\cite{gomez2008semisupervised}. LapSVM was later modified to incorporate spatial-spectral information~\cite{yang2014semi} and semi-supervised kernel propagation with sparse coding~\cite{yang2017sparse}.  Ratle et al.~\cite{ratle2010semisupervised} recognized the shortcomings of using an SVM and replaced it with a semi-supervised neural network.  This neural network outperformed LapSVM and TSVM on the Indian Pines and Kennedy Space Center HSI datasets in both classification accuracy and computational efficiency.  

Dopido et al.~\cite{dopido2014new} introduced a semi-supervised model that jointly learned the classification and spectral unmixing task to help improve classification performance on training sets with only a few labeled samples.  Liu et al.~\cite{liu2017semi} used the ladder network architecture proposed by \cite{rasmus2015semi} to semantically segment HSI.  Their ladder network model used convolutional hidden layers in order to learn spatial-spectral features directly from the image.  Both of these frameworks introduce an unsupervised task that is jointly optimized with the classification task to help regularize the model, which helped the model generalize and perform well with smaller training sets.

\section{Methods}
\label{section:methods}

\subsection{Multi-Loss Convolutional Autoencoder}

\begin{figure}[t]
    \centering
    \includegraphics[width=0.75\linewidth]{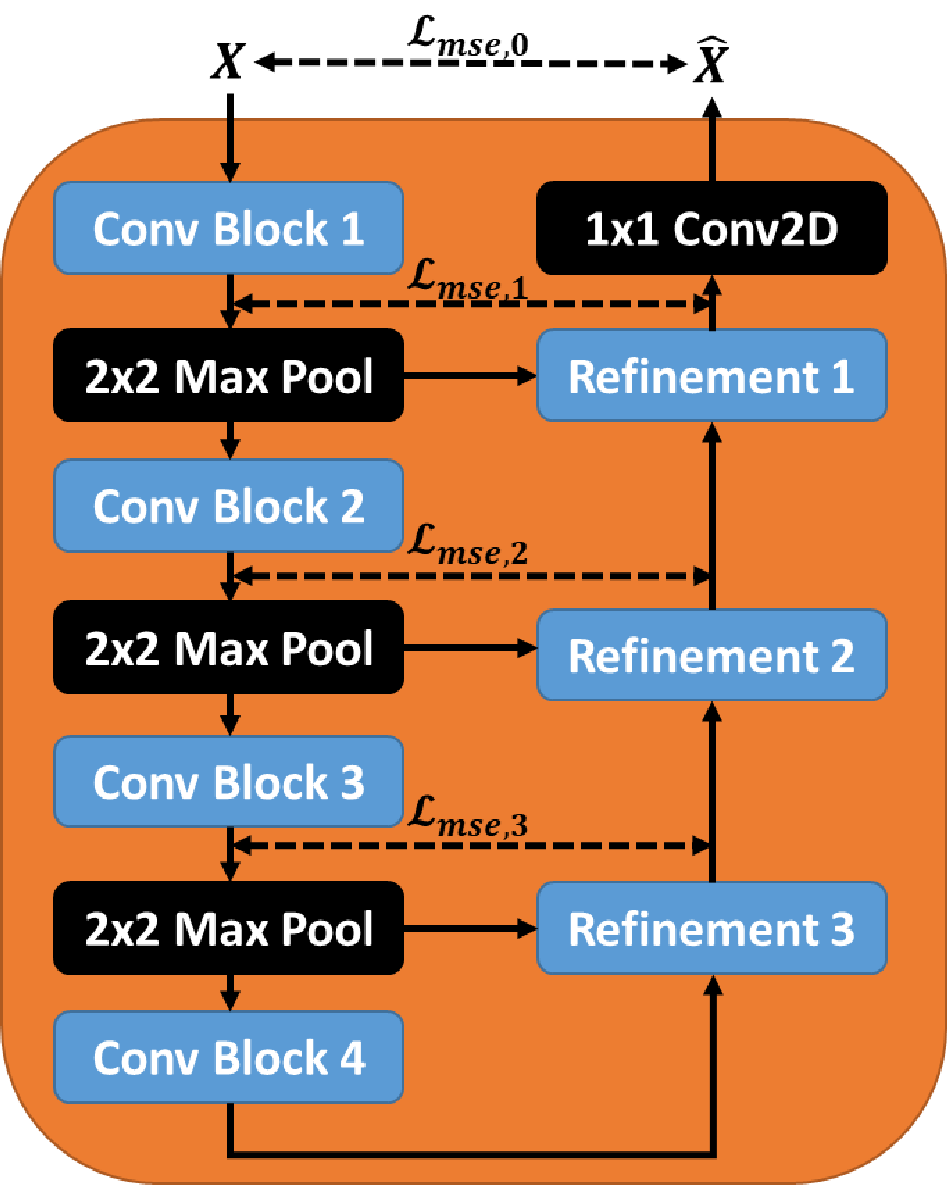}
    \caption{MCAE model architecture.  Dashed lines indicate where the mean-squared error loss $\mathcal{L}_j$ is calculated for layer $j$, and solid lines are the feed-forward and lateral network connections where information is passed.  The refinement layers (Fig.~\ref{fig:refinement} are responsible for reconstructing the downsampled feature response.}
    \label{fig:mcae}
\end{figure}

\begin{figure}[t]
    \centering
    \includegraphics[width=0.75\linewidth]{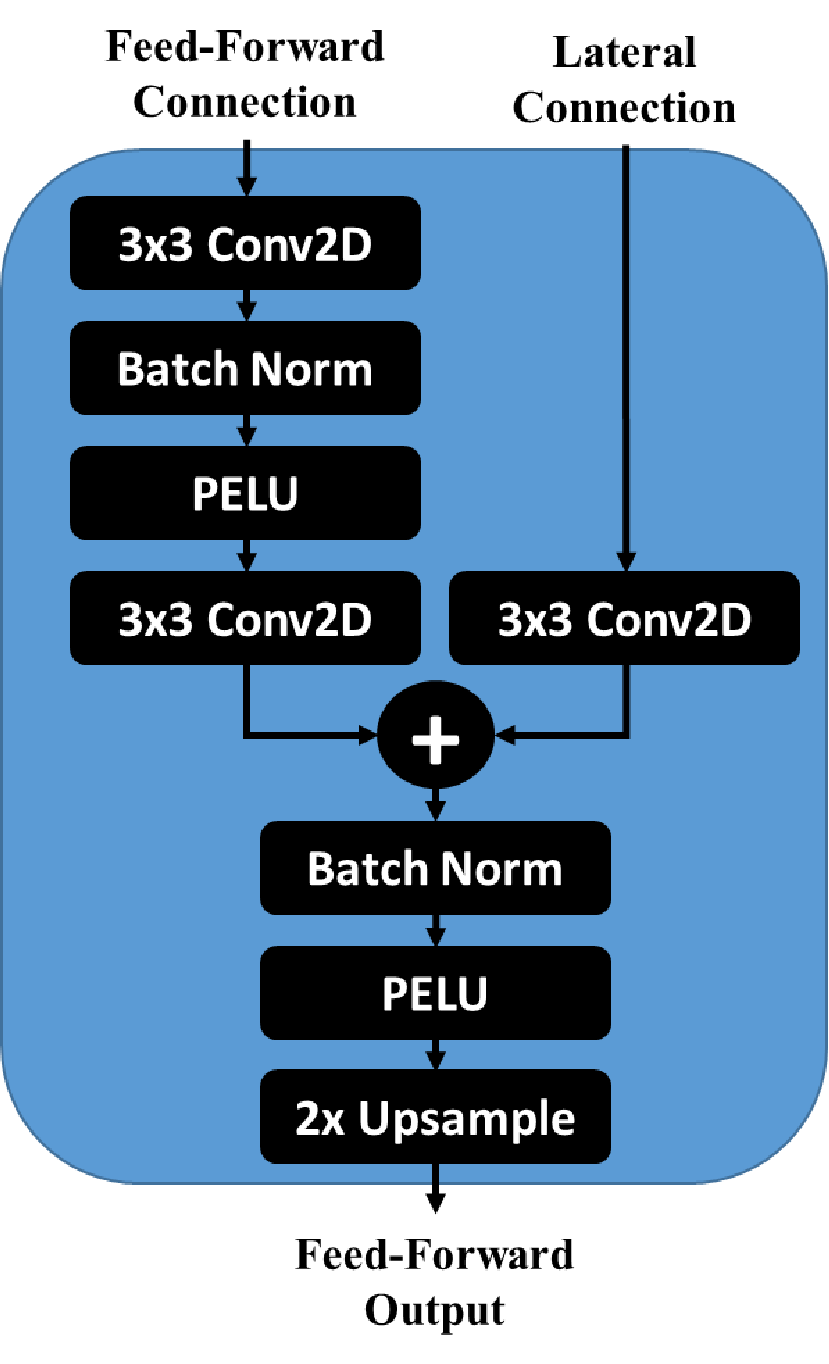}
    \caption{Refinement layer used in CAE and MCAE.}
    \label{fig:refinement}
\end{figure}

Here, we describe the MCAE model (Fig.~\ref{fig:mcae}), a significant improvement over the original CAE model~\cite{kemker2017self}. Formally, an autoencoder $f$ is an unsupervised neural network that attempts to reconstruct the input $\textbf{x}$ (e.g., sample from HSI) such that $\hat{\textbf{x}} = f(\textbf{x})$, where $\hat{\textbf{x}}$ is the autoencoder's reconstruction of the original input $\mathbf{x}$.  An autoencoder can be trained with various constraints to learn a meaningful feature representation that can still be used for reconstruction.  Typically, autoencoders include a separate encoder network that learns a compressed feature representation of the data and a symmetrical decoder network that reconstructs the compressed feature representation back into an estimate of the original input.  These networks have hidden layers that use trainable weights $\textbf{W}$ and biases $\textbf{b}$ to compress and then reconstruct the input.  Since this is an unsupervised learning method, the MSE between $\textbf{x}$ and $\hat{\textbf{x}}$ is the loss used to train the network.  Once trained, we can extract the features $\textbf{h}$ from an autoencoder with a single hidden-layer such that,
\begin{equation}
\label{eq:ae}
\textbf{h} = \sigma\left(\textbf{W}\textbf{x} + \textbf{b}\right)
\end{equation}
where $\sigma$ is the non-linear activation function (e.g., CAE used the Rectified Linear Unit (ReLU) activation).  A CAE replaces multiply/add operations with 2-D convolution operations, 

\begin{equation}
\label{eq:cae}
\textbf{H} = \sigma\left(\textbf{W} * \textbf{X}\right)
\end{equation}
where $*$ denotes the 2-D convolution operation and $\textbf{X}$ is the 2-D image data that will be convolved.  2-D convolution operations learn position invariant feature representations; that is, the feature response for a given object in an image is independent of the pixel location.  It slides learned convolution filters across the target image, so the number of trainable parameters are $k^2 \times F_{in} \times F_{out}$, where $k$ is the number of pixels along the edge of the convolution filter (e.g., typically $k=3$), and $F_{in}$ and $F_{out}$ are the number of input/output features respectively .  Standard multi-layer perceptron (MLP) neural networks have a trainable parameter relating every pixel to every input/output feature, resulting in $N_{pixels}^2 \times F_{in} \times F_{out}$ trainable parameters, where $N_{pixels}$ is the number of pixels in the image data.  DCNNs almost always have fewer trainable parameters than MLPs of equivalent depth, which can prevent the model from overfitting.  In  \cite{kemker2017self}, the stacked CAE (SCAE) model is built using several CAEs, where the input to the \(k\)-th CAE is the output from the last hidden-layer of the \(k-1\)-th CAE,
\begin{equation}
\label{eq:scae}
\textbf{H}^k = \sigma\left(\textbf{W}^k * \textbf{H}^{k-1}\right)
\end{equation}
where $\textbf{H}^0=\textbf{X}$.  This allowed the model to learn a deeper feature representation from the input data.  Each CAE contains multiple hidden-layers and the down-sampled feature response is reconstructed by the refinement layer shown in Fig.~\ref{fig:refinement}.

\begin{figure*}[t]
    \centering
    \includegraphics[width=0.9\linewidth]{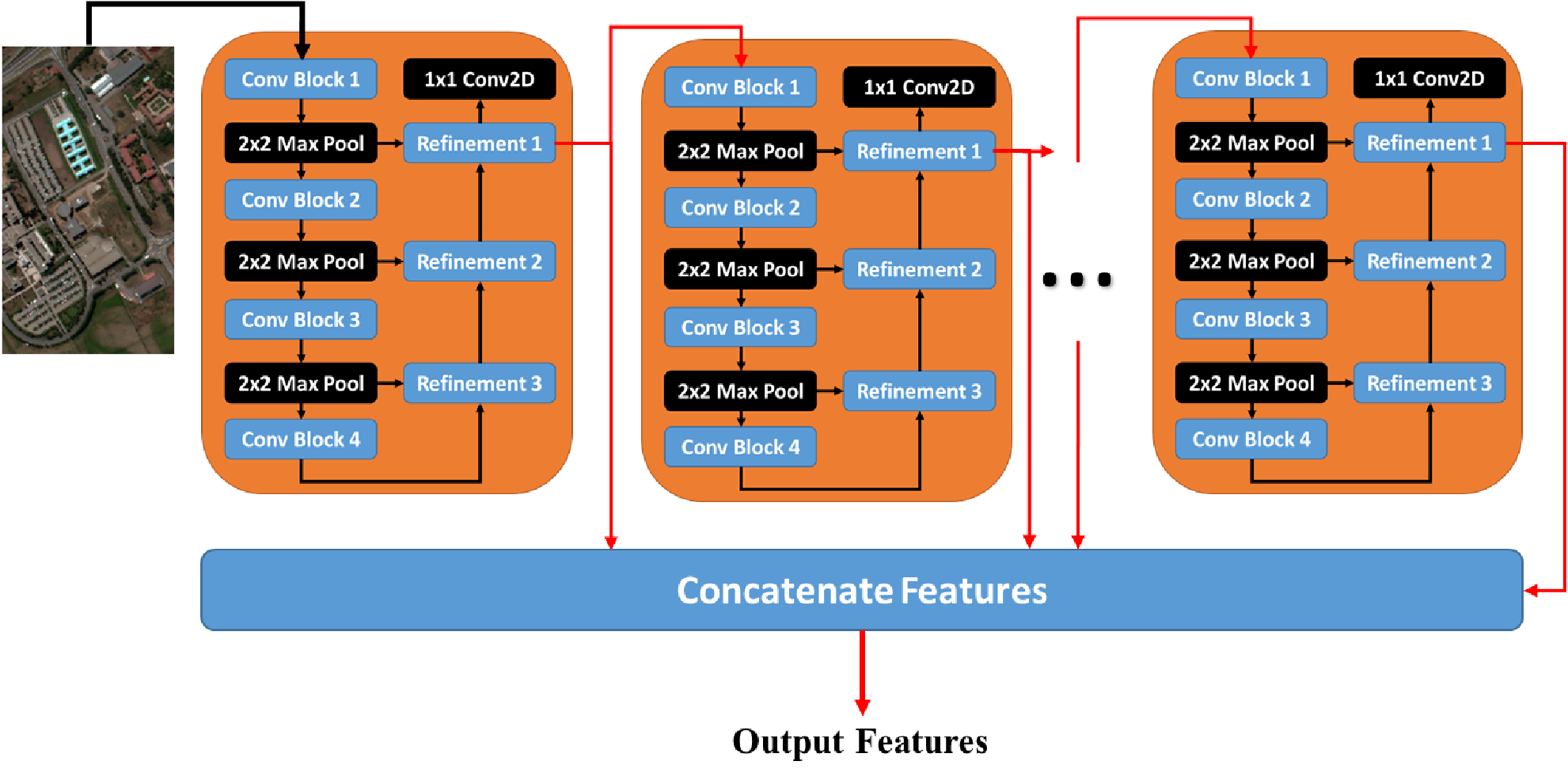}
    \caption{The stacked multi-loss convolutional autoencoder (SMCAE) spatial-spectral feature extractor used in this paper consists of two or more MCAE modules.  The red lines denote where features are being extracted, transferred to the next MCAE, and concatenated into a final feature response.}
    \label{fig:smcae_model}
\end{figure*}

Valpola~\cite{valpola2015neural} showed that, for an autoencoder with multiple hidden-layers, errors in deeper layers had a harder time being corrected during back-propagation because they are too far from the training signal.  To fix this, we train each CAE using a weighted sum of the reconstruction losses for each hidden-layer,

\begin{equation}
\label{eq:mcae}
\mathcal{L} = \sum\limits_{j=1}^M \lambda_{mcae,j} \mathcal{L}_{mse,j} 
\end{equation}
where $M$ is the number of hidden-layers, $\mathcal{L}_{mse,j}$ is the MSE of the encoder and decoder at layer $j$, and $\lambda_{mcae,j}$ is the loss weight at layer $j$.  We refer to this new feature extracting model as MCAE.  The SMCAE model (Fig. \ref{fig:smcae_model}) trains, extracts, and concatenates feature responses from the individual MCAEs in the same manner as SCAE.

\subsection{Semi-Supervised Multi-Layer Perceptron Neural Network}

In \cite{kemker2017self}, a major bottleneck in their self-taught learning model was that it used PCA to reduce the feature dimensionality prior to being classified by an SVM.  This was necessary because SVMs can suffer from the curse of dimensionality when the feature dimensionality is too high. The ideal number of principal components varied across datasets and required cross-validation. In contrast, MLP-based neural networks are able to learn what features are most important for semantic segmentation.  The downside is that standard MLPs require large quantities of labeled data or they will overfit. 

\begin{figure}[t]
    \centering
    \includegraphics[width=0.90\linewidth]{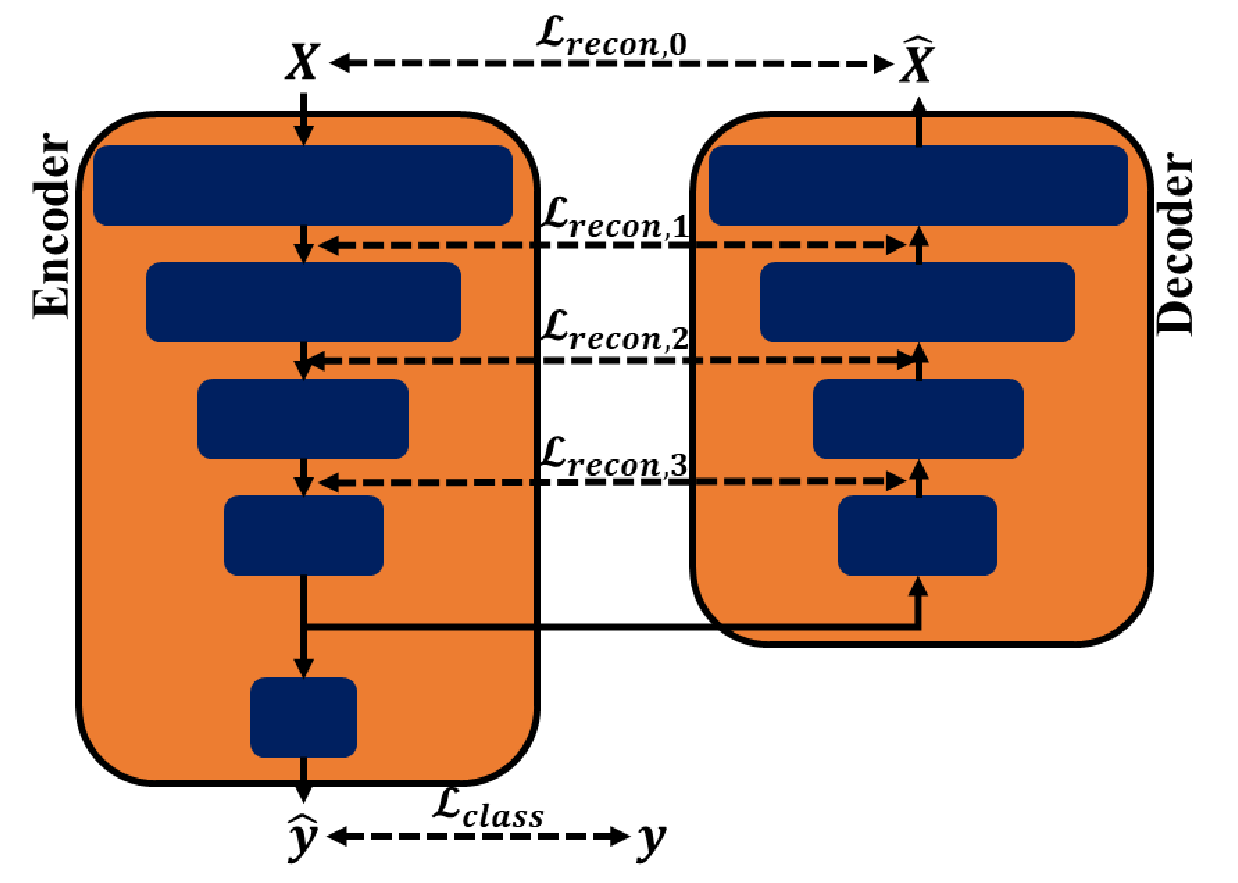}
    \caption{The semi-supervised multi-layer perceptron (SS-MLP) classification framework used in this paper.}
    \label{fig:ssmlp}
\end{figure}

To overcome this problem we propose a semi-supervised MLP (SS-MLP). As shown in Fig.~\ref{fig:ssmlp}, SS-MLP has a symmetric encoder-decoder framework.  The feed-forward encoder network segments the original input and the decoder reconstructs the compressed feature representation back to the original input.  The reconstruction serves as an additional regularization operation that can prevent the model from overfitting when there are only a few training samples available.  SS-MLP is trained by minimizing the total supervised and unsupervised loss
\begin{equation}
\mathcal{L} = \mathcal{L}_{class} + \sum\limits_{j=1}^M\lambda_{recon,j} \cdot \mathcal{L}_{recon, j}
\end{equation}
\noindent where $\mathcal{L}_{class}$ is the cross-entropy loss for classification, $\mathcal{L}_{recon,j}$ is the MSE of the reconstruction at layer $j$, $\lambda_{recon, j}$ is the importance of the unsupervised loss term at layer $j$, and $M$ is the number of hidden layers in SS-MLP.  The $\lambda_{recon, j}$ weights are set empirically.  This optimization strategy is similar to the ladder network introduced in~\cite{liu2017semi},  where the network uses convolutional units to learn spatial-spectral features from a single HSI cube.  In this case, the learned spatial-spectral features are specific to this dataset alone and may not transfer well to other HSI we wish to classify.  Self-taught learning features, which are learned from a large quantity of imagery, can be more discriminative and generalize well across multiple datasets.  In this paper, we will use pre-trained SMCAE models to extract features from the labeled data and then pass them to SS-MLP to generate the final classification map.    

\subsection{Adaptive Non-Linear Activations}

Kemker and Kanan~\cite{kemker2017self} showed that classification performance with low-level features could be improved by applying an adaptive non-linearity to the feature response.  The SCAE is a deep-feature extractor, but it only used a Rectified Linear Unit (ReLU) activation which just sets all negative values to zero.  Fixed activations like this may not be the ideal non-linearity required for every network layer; so in this paper, we use the Parametric Exponential Linear Unit (PELU) activation~\cite{trottier-pelu}, 
\begin{equation}
\sigma\left(\textbf{h}; a,b\right) = 
\begin{cases}
	\frac{a}{b} \textbf{h} & \text{if } \textbf{h} \geq 0 \\
    a\left(\text{exp}\left(\frac{\textbf{h}}{b}\right)-1\right) & \text{otherwise} 
\end{cases} 
\end{equation}
\noindent where $a$ and $b$ are positive trainable parameters.  PELU was shown to increase performance by learning the ideal activation function for each network layer~\cite{trottier-pelu}.  Depending on the values of $a$ and $b$, PELU can approximate a ReLU activation function or a number of other commonly used activation functions (e.g., LeakyReLU~\cite{maas2013rectifier} and exponential linear units~\cite{clevert2015fast}).  In this paper, we use PELU activations with our SMCAE feature extractor and SS-MLP classifier.

\section{Experimental Setup}

\subsection{Data Description}

\begin{figure}[th!]
  \centering
  \subfigure[AVIRIS]{%
  \includegraphics[height=2.7cm]{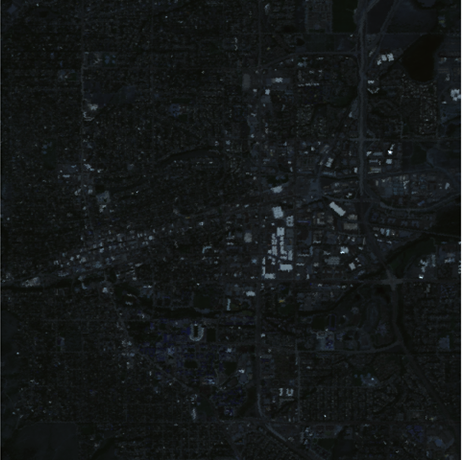}
  \label{fig:aviris}}
  \subfigure[Hyperion]{%
  \includegraphics[height=2.7cm]{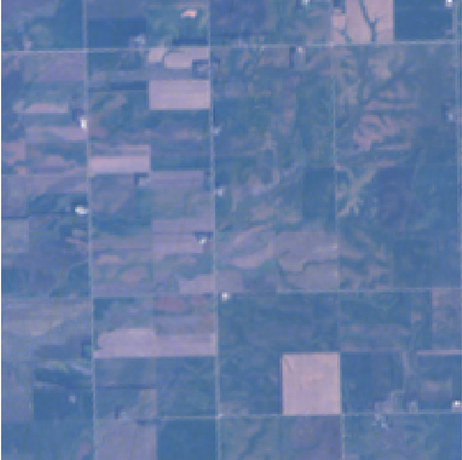}
  \label{fig:hyperion}} 
  \subfigure[GLiHT]{%
  \includegraphics[height=2.7cm]{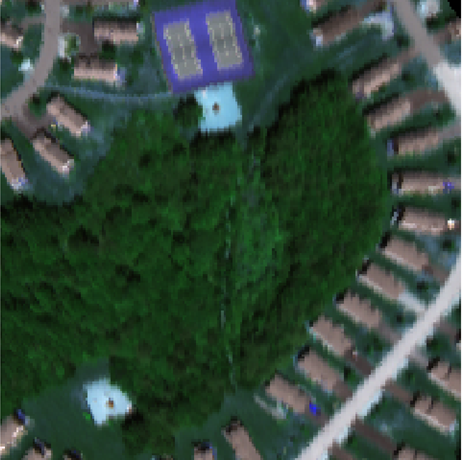}
  \label{fig:gliht}} 
  \caption{RGB visualization of HSI from all three sensors used to train SCAE and SMCAE.}
  \label{fig:samples}
\end{figure}

The SCAE and SMCAE frameworks were trained using publicly-available HSI data collected by three different NASA sensors:  1) NASA Jet Propulsion Laboratory's Airborne Visible/Infrared Imaging Spectrometer (AVIRIS), 2) EO-1 Hyperion imaging spectrometer, and 3) Goddard's LiDAR, Hyperspectral \& Thermal Imager (GLiHT).  Relevant technical specifications for each sensor are available in Table~\ref{table:hsi_unlabeled}. We attempted to collect data from a wide variety of different locations and climates (e.g., urban, forest, farmland, etc.) so that the frameworks would learn spatial-spectral features that generalize across multiple labeled datasets.  Samples from all three sensors can be seen in Fig.~\ref{fig:samples}.

\begin{table}[t!]
\centering \footnotesize
\caption{Various HSI sensors used in this paper to train and evaluate our SMCAE SS-MLP framework.}
\begin{tabular}{@{}l|cccc@{}}\toprule
& \textbf{AVIRIS} & \textbf{Hyperion} & \textbf{GLiHT} & \textbf{ROSIS} \\
\midrule
\textbf{Platform} & Airborne & Satellite  & Airborne & Airborne \\
\textbf{Spectral} & \multirow{2}{*}{400-2500} & \multirow{2}{*}{400-2500}  & \multirow{2}{*}{400-1000} & \multirow{2}{*}{430-838} \\
\textbf{Range [nm]} \\
\textbf{Spectral} & \multirow{2}{*}{224} & \multirow{2}{*}{220} & \multirow{2}{*}{402} & \multirow{2}{*}{115}\\
\textbf{Bands [\#]}\\
\textbf{FWHM [nm]} & 10 & 10 & 5 & 5 \\
\textbf{GSD [m]} & Varies & 30 & $<1$ & 0.3-0.7 (best) \\
\multirow{2}{*}{\textbf{Sensor Type}} & Whisk  & Grating Image & 2-D CCD & Grating Image \\
 &Broom &Spectrometer & Imager &Spectrometer\\
 \midrule
 \multicolumn{5}{l}{AVIRIS - Airborne Visible/Infrared Imaging Spectrometer}\\
  \multicolumn{5}{l}{CCD - Charged Couple Device} \\
 \multicolumn{5}{l}{FWHM - Full-width, Half-Max}\\
 \multicolumn{5}{l}{GSD - Ground Sample Distance} \\
 \multicolumn{5}{l}{GLiHT - Goddard's LiDAR, Hyperspectral \& Thermal Imager}\\
 \multicolumn{5}{l}{ROSIS - Reflective Optics System Imaging Spectrometer}\\
\bottomrule 
\end{tabular}
\label{table:hsi_unlabeled}
\end{table}

\begin{figure}[th!]
  \centering
  \subfigure[Indian Pines]{%
  \includegraphics[height=4cm]{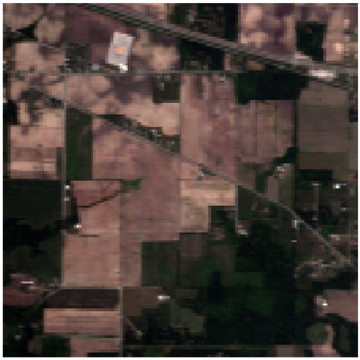}   	       \label{fig:ip}}
  \subfigure[Pavia Univ.]{%
  \includegraphics[height=4cm]{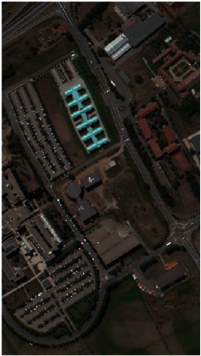}
  \label{fig:pavia}} 
  \subfigure[Salinas]{%
  \includegraphics[height=4cm]{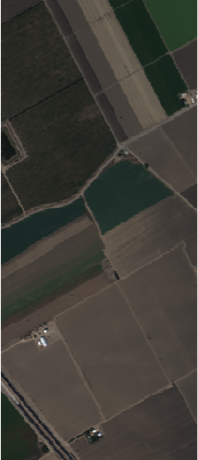}
  \label{fig:salinas}} 
  \caption{RGB visualization for Indian Pines, Pavia University, and Salinas Valley HSI datasets.  See Table~\ref{table:hsi} for scale.}
  \label{fig:rgb_vis}
\end{figure}

\begin{figure}[th!]
  \centering
  \subfigure[Indian Pines]{%
  \includegraphics[height=4cm]{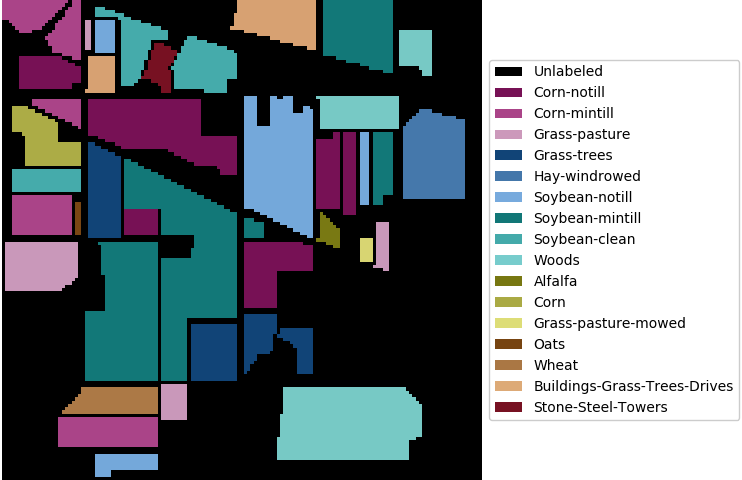}}
  \subfigure[Pavia Univ.]{%
  \includegraphics[height=4cm]{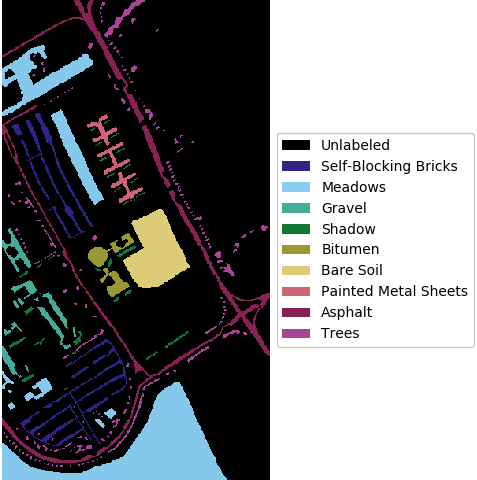}} 
  \subfigure[Salinas]{%
  \includegraphics[height=4cm]{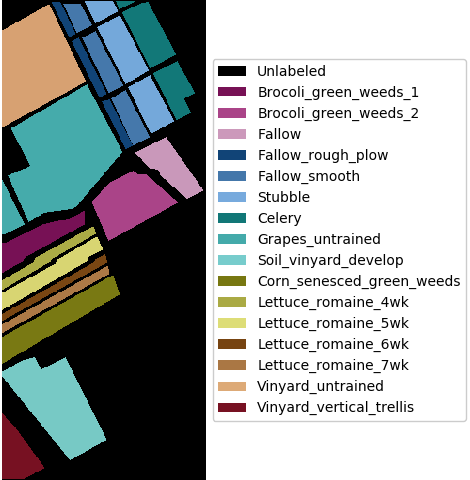}} 
  \caption{Classification truth maps for Indian Pines, Pavia University, and Salinas Valley HSI datasets.}
  \label{fig:truth_maps}
\end{figure}

The three annotated HSI datasets used to evaluate the SuSA framework are Indian Pines (Fig.~\ref{fig:ip}), Pavia University (Fig.~\ref{fig:pavia}), and Salinas Valley (Fig.~\ref{fig:salinas}).  Indian Pines and Salinas Valley were captured by the AVIRIS HSI sensor and contain mostly agricultural scenes.  Pavia University was collected by the Reflective Optics System Imaging Spectrometer (ROSIS) airborne sensor and is an urban scene with several man-made objects.  Fig.~\ref{fig:rgb_vis} shows a RGB visualization of all three datasets and Fig.~\ref{fig:truth_maps} shows their corresponding ground truth maps.

\begin{table}[t!]
\centering \footnotesize
\caption{Benchmark HSI datasets used in this paper to evaluate the algorithms. }
\begin{tabular}{@{}rccc@{}}\toprule
 & \textbf{Indian} & \textbf{Pavia} & \textbf{Salinas} \\
 & \textbf{Pines} & \textbf{University} & \textbf{Valley}\\
 \midrule
 \textbf{Sensor} & AVIRIS & ROSIS & AVIRIS\\
 \textbf{Spatial Dimensions [pix]} & $145\times145$ & $610\times340$ & $512\times217$\\
 \textbf{GSD [m]} &20 & 1.3 & 3.7 \\
 \textbf{Spectral Bands} & 224 & 103 & 224\\
 \textbf{Spectral Range [nm]}  & 400-2500 & 430-838 & 400-2500 \\
 \textbf{Number of Classes} & 16 & 9 & 16\\
 \midrule
 \multicolumn{4}{l}{GSD - Ground Sample Distance} \\
 \multicolumn{4}{l}{ROSIS - Reflective Optics System Imaging Spectrometer}\\
 \multicolumn{4}{l}{AVIRIS - Airborne Visible/Infrared Imaging Spectrometer}\\
\bottomrule \\
\end{tabular}
\label{table:hsi}
\end{table}

\subsection{Training Parameters}

\subsubsection{MCAE}

The CAE and MCAE frameworks use the parameters listed in Table \ref{table:cae_params} throughout this paper.  We used the same layer shape found to work well in~\cite{kemker2017self} for CAE and MCAE to provide a fair comparison between the two models.  These networks were trained using the open-source imagery listed in Table \ref{table:hsi_unlabeled}.  We randomly sampled a total of 50,000 32$\times$32$\times B$ image patches from these different HSI images, where $B$ is the number of spectral bands that correspond to each sensor.  Of the 50,000 image patches, 45,000 are reserved for training and 5,000 are reserved for validation.  Bands that correspond to low SNR and atmospheric absorption are removed.  We center each feature in the patch array to zero-mean and unit-variance prior to training the model.  The weights are initialized with Xavier initialization~\cite{glorot2010understanding}(i.e., drawn from a normal distribution with its variance chosen based on the number of units), and the biases and PELU parameters are initialized with ones.

\begin{table}[t]
\centering \footnotesize
\caption{Training parameters for CAE and MCAE.}
\begin{tabular}{@{}l|c|c@{}}\toprule
& \textbf{CAE} & \textbf{MCAE} \\
\midrule 
\textbf{Multi-Loss Weights} & None & 1, $10^{-1}$, $10^{-2}$, $10^{-2}$ \\
\textbf{Convolution Layer} & \multicolumn{2}{c}{256,512,512,1024} \\
\textbf{Refinement Layer} & \multicolumn{2}{c}{512,512,256}\\
\textbf{Activation} &  ReLU & PELU \\
\textbf{Initial Learning Rate} &  \multicolumn{2}{c}{$2\times 10^{-3}$} \\
\textbf{Batch Size} &  \multicolumn{2}{c}{512} \\
\bottomrule 
\end{tabular}
\label{table:cae_params}
\end{table}

SCAE and SMCAE were trained using the Nadam optimizer, which is a common variant of stochastic gradient descent used to speed up training of deep learning models~\cite{dozat2016incorporating}.  During training, the learning rate was dropped by a factor of 10 when the validation loss did not improve for five consecutive epochs. The models were also trained using early stopping, where training terminated when the validation loss did not improve for ten consecutive epochs.  The output of the last hidden layer is then fed to the next CAE/MCAE to build the corresponding SCAE/SMCAE frameworks.

After training SCAE and SMCAE, we use them to extract features from the annotated datasets.  First, we re-sample the data to match the same spectral-bands and full-width, half-maxes (FWHMs) as the data used to train the corresponding feature extracting framework. Throughout this paper, we use the band resampling method used in~\cite{spectral}, which has been made publically available.  This method assumes that the target sensor has a (per-band) Gaussian response.  For each target band and corresponding full-width, half-max (FWHM), the algorithm searches for the source bands that overlap and then integrates those responses over the region of overlap in the target sensor. 

Next, we center each feature in the data to zero-mean/unit-variance.  Finally, we pass this data through the first CAE/MCAE and extract the features from the last hidden layer.  These features are fed to the second CAE/MCAE, and so on.  The output from each CAE/MCAE is concatenated along the feature dimension.  Each feature in the feature response is centered to zero-mean, unit-variance.  Finally, we incorporate translation invariance into our final feature response by pooling the feature response with a $5 \times 5$ mean-pooling filter.  The receptive field of this filter is considerably smaller than the one used in~\cite{kemker2017self}, which will prevent the mean-pooling operation from blurring out small objects and will also preserve sharp boundaries between object classes.   

\subsubsection{SS-MLP}

The input to the SS-MLP classifier is the extracted features from SMCAE.  The HSI cube is reshaped into a 2-dimensional vector (i.e., number of pixels $\times$ number of features).  The parameters for the SS-MLP classifier used in this paper are shown in Table~\ref{table:ssmlp_params}.  The relatively high weight decay term was shown in \cite{trottier-pelu} to work well for the PELU activation.  

\begin{table}[t]
\centering \footnotesize
\caption{Training parameters for SS-MLP.}
\begin{tabular}{@{}r|c@{}}\toprule
\textbf{Hidden Layer Shapes} & $\left[1600, 950, 250, 225\right]$ \\
\textbf{Activation} & PELU \\
$\boldsymbol{\lambda_{recon}}$ & $\left[1,1,0.1,0.1,0.1,0.1\right]$\\
\textbf{Initial Learning Rate} &$2\cdot 10^{-3}$\\
\textbf{Mini-Batch Size} & 8 \\
\textbf{Weight Decay} & $10^{-3}$ \\
\bottomrule 
\end{tabular}
\label{table:ssmlp_params}
\end{table}

The weights are initialized with Xavier initialization~\cite{glorot2010understanding}, and the biases and PELU parameters are initialized with ones.  We optimize the joint loss function using the Nadam optimizer.  The initial learning rate is the default $2\cdot 10^{-3}$.  We drop the learning rate by a factor of 10 when the validation accuracy plateaus for 25 consecutive epochs; and we stop training the model when the validation accuracy plateaus for 50 consecutive epochs.  The training/validation folds are built by randomly sampling the available training data 90\%/10\% respectively. 

\section{Experimental Results and Discussion}

We conducted experiments to measure the performance of our proposed SuSA framework. All of the results are reported as the mean and standard deviation of 30 trials. In each trial, we randomly sample \(L\) labeled samples from the HSI dataset for training. The reported performance is the semantic segmentation result on all available labeled samples.  The three reported metrics used for this section are overall accuracy (OA), mean-class (average) accuracy (AA), and Cohen's kappa coefficient ($\kappa$).

Before giving the results of the full model across three datasets in Section~\ref{section:fusion}, we first describe preliminary experiments to compare single- vs. multi-loss CAE and study the effect of stacking features using the Pavia University dataset. 

\subsection{Single- vs. Multi-Loss CAE}
\label{section:firstexp}

In this section, we compare the CAE model proposed earlier in \cite{kemker2017self} to the MCAE model proposed in this paper using the Pavia University dataset for both $L=10$ and $L=50$ samples per class.  In this experiment, we extracted the features from a single CAE/MCAE trained on unlabeled AVIRIS HSI. The results are given in Table~\ref{table:cae}. We also show performance on the raw spectrum (i.e., pass the original HSI to SS-MLP).  MCAE outperforms its CAE predecessor, although the gap is not large. In the next sections, we increase this gap by including features from stacked MCAEs trained by HSI from three different sensors.  

\begin{table}[t]
\centering \footnotesize
\caption{Classification results on the Pavia University dataset using a single CAE and MCAE model trained on unlabeled AVIRIS data.  These results were generated by training SS-MLP on \(L\) labeled samples per class. Best performance for each experiment is in bold.}
\begin{tabular}{@{}l|ccc@{}}\toprule
 & \textbf{OA} & \textbf{AA} & \boldsymbol{$\kappa$}\\
 \midrule
 \text{\(\bm{L=10}\)} \\
 \text{\textbf{Raw Spectrum}} & $77.58\pm2.41$ & $76.37\pm3.31$ & $0.7041\pm0.0306$  \\
 \text{\textbf{CAE}} & $83.47\pm2.66$ & $84.78\pm2.99$ & $0.7844\pm0.0325$ \\
 \text{\textbf{MCAE}} & $\textbf{84.07}\pm\textbf{2.49}$ &  $\textbf{84.95}\pm\textbf{2.54}$ & $0.\textbf{7923}\pm\textbf{0.0305}$\\[1ex]
 \(\bm{L=50}\) \\
 \text{\textbf{Raw Spectrum}} & $87.93\pm0.92$ & $87.55\pm1.15$ & $0.8411\pm0.0117$  \\
 \text{\textbf{CAE}} & $92.08\pm1.23$ & $92.53\pm1.55$ & $0.8960\pm0.0158$ \\
 \text{\textbf{MCAE}} & $\textbf{94.19}\pm\textbf{0.99}$ &  $\textbf{94.74}\pm\textbf{1.16}$ & $\textbf{0.9234}\pm\textbf{0.0130}$ \\
\bottomrule 
\end{tabular}
\label{table:cae}
\end{table}

\subsection{Stacked Feature Representations}
In this experiment, we examine the impact of stacking MCAE feature representations on classification performance.  We extracted features from the SMCAE model, trained on unlabeled AVIRIS HSI, and fed it to four different classifiers: linear kernel SVM, radial basis function (RBF) SVM, standard MLP, and our SS-MLP.  For the SVM experiments, we cross-validate for the optimal cost $C$ and kernel width $\gamma$ (RBF only) hyperparameters.  We use the same hyperparameters in Table \ref{table:ssmlp_params} for the standard and semi-supervised MLP classifiers. 

\begin{figure}[t]
    \centering
    \includegraphics[width=0.90\linewidth]{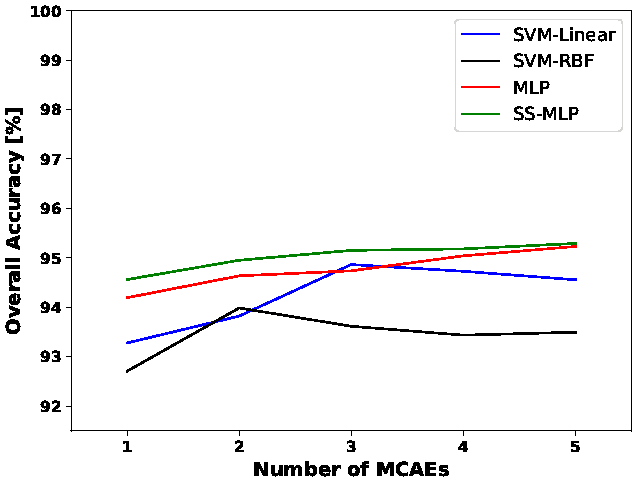}
    \caption{SMCAE performance on four different classifiers: linear kernel SVM, radial basis function (RBF) SVM, standard MLP, and our SS-MLP.  Our SS-MLP model does the best.}
    \label{fig:smcae}
\end{figure}

Each model was trained on Pavia University using $L=50$ samples per class.  Fig.~\ref{fig:smcae} shows the  mean-class test accuracy of each classifier (as a mean of 30 runs) as additional stacked MCAE feature representations are added.   SS-MLP model outperformed these standard classification methods and  the performance improves as more MCAE features are added.  Since the performance saturates at 4-5 MCAEs, we will use 5 MCAEs from each sensor for the remainder of this paper.  The SVM classifier's peak performance occurs at 2-3 CAEs and then decreases when additional CAE features are added due to overfitting.  

\subsection{Multi-Sensor Fusion}
\label{section:fusion}
In this section, we show how combining features from SMCAE models trained on HSI collected from different sensors can significantly improve semantic segmentation performance.  In this experiment, we evaluate performance using the Pavia University dataset, where our framework is trained using $L=50$ samples per class.  We trained three variants of SMCAE, where the model is trained on HSI from the AVIRIS, Hyperion, and GLiHT sensors.  We also tested each possible combination of SMCAE frameworks, where the output of each SMCAE is concatenated along the feature axis.  Table~\ref{table:fusion} shows the impact that each SMCAE has on performance.  Performance across models differs noticeably, and combining features from multiple sensors yields the best performance.  This could indicate that each SMCAE model learns novel information that is not available from the SMCAE models trained on different sensors (see Section~\ref{section:discussion} for more details).  The SMCAE model trained on GLiHT yielded superior results than the other two SMCAE models.  This is likely because Pavia University and GLiHT share similar spectral range and bands; whereas AVIRIS and Hyperion expand beyond the range covered by the ROSIS sensor that collected Pavia University.  

\begin{table}[t]
\centering \footnotesize
\caption{Classification performance using features extracted from SMCAE models that were trained with data from different HSI sensors.}
\begin{tabular}{@{}c|ccc@{}}\toprule
 \textbf{Data Source(s)} & \textbf{OA} & \textbf{AA} & \boldsymbol{$\kappa$} \\
 \midrule
 \textbf{AVIRIS}   & $95.07\pm0.69$ &$96.03\pm0.85$ & $0.9350\pm0.0090$\\
 \textbf{Hyperion} & $95.93\pm0.90$& $96.47\pm0.56$ & $0.9463\pm0.0117$\\
 \textbf{GLiHT}    & $97.83\pm0.67$& $98.03\pm0.57$&$0.9713\pm0.0088$\\
 \textbf{AVIRIS/Hyperion} &$96.51\pm0.91$ &$96.85\pm1.06$ & $0.9538\pm0.0120$\\
 \textbf{AVIRIS/GLIHT}    &$97.96\pm0.57$ &$98.13\pm0.49$ &$0.9730\pm0.0075$\\
 \textbf{Hyperion/GLiHT}  &$98.13\pm0.36$ &$98.21\pm0.38$ & $0.9752\pm0.0048$\\
 \textbf{AVIRIS/Hyperion/} & \multirow{2}{*}{$\textbf{98.18}\pm\textbf{0.53}$} & \multirow{2}{*}{$\textbf{98.29}\pm\textbf{0.38}$} & \multirow{2}{*}{$\textbf{0.9759}\pm\textbf{0.0069}$}\\
 \textbf{GLiHT} \\
\bottomrule 
\end{tabular}

\label{table:fusion}
\end{table}

\subsection{State-of-the-Art Comparison}
\label{section:stateofart}
In this section, we use the same SMCAE configuration discussed in Section \ref{section:fusion}, where we stacked features from all three sensors listed in Table~\ref{table:fusion}.  Table~\ref{table:10samples} shows the classification performance when $L=10$ samples per class.  We compared against models found to work well using this training paradigm.  For Indian Pines and Pavia University, we compare against a semi-supervised classification approach that uses spectral-unmixing to help improve classification performance~\cite{dopido2014new}.  They showed that introducing the unsupervised task helped regularize the model, thus improving generalization when only small quantities of annotated image data are available.  Their results were reported as the mean and standard deviation of 10 separate runs.  Imani~and~Ghassemian~\cite{imani2013boundary} proposed a model that was supposed to work well on all three of the annotated HSI datasets evaluated in this paper; however, they showed that a SVM classifier yielded the best results.  They only reported the mean (no standard deviation) of the mean-class accuracy over three runs.  To generate more detailed results, we reproduced this experiment using an SVM-RBF classifier.  We reported the overall accuracy, mean-class accuracy, and kappa statistic as the mean and standard deviation over 30 trials.  Our SuSA framework achieved superior results compared to each of these frameworks.
 
\begin{table}[t]
\centering \footnotesize
\caption{Results of low-shot learning experiment where the training set contains only \(L\)=10 samples per class.}
\begin{tabular}{@{}lccc@{}}\toprule
 \textbf{Model} & \textbf{OA} & \textbf{AA} & \boldsymbol{$\kappa$}  \\
\midrule
\multicolumn{1}{@{}l}{\textbf{Indian Pines}} \\
 Dopido et al.~\cite{dopido2014new}&  $75.29\pm2.40$ & $79.05\pm2.00$ & $0.7184\pm0.0275$\\
 MICA-SVM~\cite{kemker2017self} & $66.32\pm2.17$  &$81.47\pm1.27$ &$0.6261\pm0.0237$ \\
SCAE-SVM~\cite{kemker2017self} & $80.58\pm2.13$  &$89.24\pm1.08$ &$0.7816\pm0.0234$ \\
 SuSA & $\textbf{81.16}\pm\textbf{1.85}$ &   $\textbf{89.01}\pm\textbf{1.26}$ & $\textbf{0.7874}\pm\textbf{0.0205}$ \\[1ex]
\multicolumn{1}{@{}l}{\textbf{Pavia University}}\\
 Dopido et al.~\cite{dopido2014new}&  $84.14\pm1.97$ & $84.48\pm1.04$ & $0.7923\pm0.0237$\\
 MICA-SVM~\cite{kemker2017self} & $75.74\pm3.81$  &$78.85\pm4.01$ &$0.6907\pm0.0446$ \\
SCAE-SVM~\cite{kemker2017self} & $84.67\pm3.36$  &$87.32\pm2.04$ &$0.8028\pm0.0405$ \\
 SuSA & $\textbf{89.28}\pm\textbf{2.64}$ & $\textbf{89.58}\pm\textbf{1.62}$ & $\textbf{0.8595}\pm\textbf{0.0330}$\\[1ex] 
\multicolumn{1}{@{}l}{\textbf{Salinas Valley}} \\
SVM-RBF~\cite{imani2013boundary} &$82.65\pm1.49$ & $90.01\pm0.92$ & $0.8075\pm0.0165$\\
MICA-SVM~\cite{kemker2017self} & $90.08\pm1.50$  &$94.17\pm0.99$ &$0.8899\pm0.0165$ \\
SCAE-SVM~\cite{kemker2017self} & $92.74\pm1.42$  &$95.56\pm0.80$ &$0.9193\pm0.0157$ \\
 SuSA & $\textbf{93.47}\pm\textbf{1.27}$& $\textbf{96.46}\pm\textbf{0.71}$ &$\textbf{0.9274}\pm\textbf{0.0142}$\\
\bottomrule
\end{tabular}
\label{table:10samples}
\end{table}

Table~\ref{table:compare} directly compares against previous self-taught and semi-supervised frameworks discussed in this paper.  The SCAE-SVM framework introduced by~\cite{kemker2017self} performed well on the $L=50$ samples per class training paradigms.  Note, the Indian Pines dataset used $L=50$ samples per class except for the three classes that had the smallest number of annotated training samples available, where we only used $L=15$ samples per class.  Liu~et~al.~\cite{liu2017semi} only evaluated their ladder network on Pavia University with $L=200$ samples per class.  In every case, SuSA outperforms the previous state-of-the-art classification frameworks. 

\begin{table}[t]
\centering \footnotesize
\caption{Performance comparison of SuSA against the other semi-supervised and self-taught learning frameworks discussed in this paper.}
\begin{tabular}{@{}lccc@{}}\toprule
  \textbf{Model} & \textbf{OA} & \textbf{AA} & \boldsymbol{$\kappa$}  \\
\midrule
\multicolumn{4}{@{}l}{\textbf{Indian Pines} $\left(\textbf{L}=\textbf{50/15}\right)$}\\
DAFE~\cite{ghamisi2014automatic} & 93.27 & 95.86 & 0.923\\
MICA-SVM~\cite{kemker2017self} & $94.63\pm1.00$ & $97.31\pm0.37$ & $0.9385\pm0.0114$ \\
SCAE-SVM~\cite{kemker2017self} &  $96.12\pm0.78$ & $94.58\pm0.31$ & $0.9554\pm0.0078$\\
SuSA & $\textbf{96.49}\pm\textbf{0.69}$ & $\textbf{98.34}\pm\textbf{0.31}$ & $\textbf{0.9602}\pm\textbf{0.0089}$ \\[1ex]
\multicolumn{4}{@{}l}{\textbf{Pavia University} $\left(\textbf{L}=\textbf{50}\right)$} \\
SSAE~\cite{tao2015unsupervised}& $91.96\pm0.87$ & $93.52\pm0.42$ & $0.9025\pm0.0112$ \\
MICA-SVM~\cite{kemker2017self} & $93.92\pm1.38$  &$95.58\pm0.64$ &$0.9203\pm0.0177$ \\
SCAE-SVM~\cite{kemker2017self} & $95.84\pm0.94$ & $96.56\pm0.51$ & $0.9451\pm0.0123$ \\
SuSA & $\textbf{98.18}\pm\textbf{0.53}$ & $\textbf{98.29}\pm\textbf{0.38}$ &$\textbf{0.9759}\pm\textbf{0.0069}$\\[1ex]

\multicolumn{4}{@{}l}{\textbf{Pavia University} $\left(\textbf{L}=\textbf{200}\right)$} \\
SS-CNN~\cite{liu2017semi} & 98.32 & 98.47 & Unknown \\
MICA-SVM~\cite{kemker2017self} & $98.20\pm0.33$  &$98.96\pm0.17$ &$0.9763\pm0.0043$ \\
SCAE-SVM~\cite{kemker2017self} & $98.57\pm0.24$  &$99.07\pm0.17$ &0.9812$\pm0.0032$ \\
SuSA & $\textbf{99.66}\pm\textbf{0.11}$ & $\textbf{99.70}\pm\textbf{0.09}$ & $\textbf{0.9954}\pm\textbf{0.0014}$ \\[1ex]

\multicolumn{4}{@{}l}{\textbf{Salinas Valley} $\left(\textbf{L}=\textbf{50}\right)$} \\
GLCM+~\cite{mirzapour2015improving}& $95.41$ & Unknown & Unknown \\
MICA-SVM~\cite{kemker2017self} & $97.15\pm0.56$  &$98.57\pm0.29$ &$0.9683\pm0.0062$ \\
SCAE-SVM~\cite{kemker2017self} & $98.06\pm0.45$ & $98.94\pm0.22$ & $0.9784\pm0.0050$ \\
SuSA & $\textbf{98.10}\pm\textbf{0.61}$  & $\textbf{99.11}\pm\textbf{0.26}$  & $\textbf{0.9788}\pm\textbf{0.0068}$ \\
\bottomrule
\end{tabular}

\label{table:compare}
\end{table}

We performed a statistical significance test (using a 99\% confidence interval) on the mean-class accuracy results in Table~\ref{table:compare}.  We chose mean-class accuracy because the class distributions are imbalanced, so this is a more meaningful measurement of model performance.  The results for all four training/testing paradigms were shown to be statistically significant.

\begin{figure}[th!]
  \centering
  \subfigure[Indian Pines]{%
  \includegraphics[height=3cm]{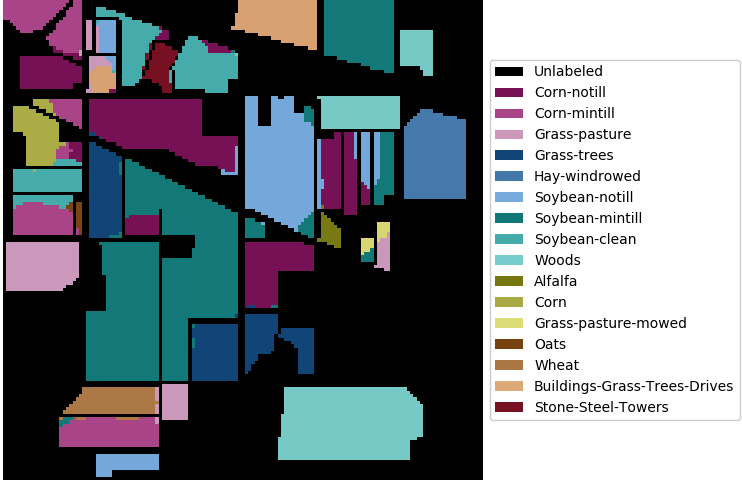}
  \label{fig:class_map_ip}}
  \subfigure[Pavia University]{%
  \includegraphics[height=3cm]{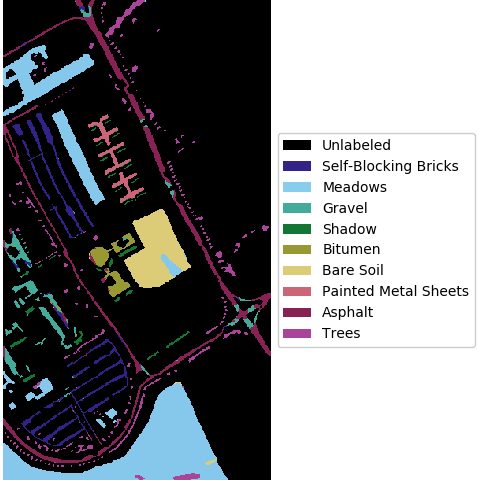}
  \label{fig:class_map_pavia}} 
  \caption{Classification maps for SuSA on the Indian Pines and Pavia University datasets from the IEEE GRSS Data and Algorithm Standard Evaluation website.}
  \label{fig:state_of_art}
\end{figure}

Finally, Table~\ref{table:state-of-art} shows that SuSA yielded state-of-the-art performance on the Indian Pines and Pavia University HSI datasets hosted on the IEEE GRSS Data and Algorithm Standard Evaluation (DASE) website.  The training/testing folds are pre-defined, and the server provides the classification performance on the test set.  This dataset is more difficult to perform well on because the training samples are co-located instead of being randomly sampled across the image. At this time, the server only lists the top-10 performers, so we are unable to ascertain the identity of the previous state-of-the-art performer or what method they used.  It also only lists their overall accuracy; however, we have provided all of the relevant statistics, including the classification maps in Fig.~\ref{fig:state_of_art}.  The main performance degradation for Pavia University occurred when SuSA predicted meadows (largest object class) when it should have predicted bare soil.  There was also a problem predicting trees when it should have predicted meadows.  For Indian Pines, SuSA mis-predicted corn for corn no-till and corn-min, and pasture/mowed grass was confused for soybeans-min.  

\begin{table}[t]
\centering \footnotesize
\caption{Classification results for the Indian Pines and Pavia University datasets from the IEEE GRSS Data and Algorithm Standard Evaluation website.}
\begin{tabular}{@{}r|cc@{}}\toprule
 & \textbf{Indian Pines} & \textbf{Pavia University}  \\
\midrule
\multicolumn{1}{l|}{\textbf{State-of-Art Performer: }} \\
\textbf{OA} & 90.73 & 73.06\\ [1ex]
\multicolumn{1}{l|}{\textbf{SuSA: }} \\
\textbf{OA} & \textbf{91.32} &\textbf{81.86} \\
\textbf{AA} & 81.17 &74.09\\
\boldsymbol{$\kappa$} & 0.90 &0.77 \\
\bottomrule 
\end{tabular}
\label{table:state-of-art}
\end{table}

\subsection{Dissimilarity Between Learned Features}
\label{section:discussion}

In this paper, we show that our SuSA framework yields state-of-the-art performance when only a few training samples are available.  We also show that transferring spatial-spectral features from multiple sensors can improve classification performance.  This would mean that SMCAE learns different features from different data and sensor modalities.  To quantify the dissimilarity between different SMCAE models, we used the dissimilarity metric proposed by \cite{kriegeskorte2008representational}.  The authors computed the dissimilarity of two feature representations $\textbf{X},\textbf{Y}$ such that,

\begin{equation}
\label{eq:dissimilarity}
d\left(\textbf{X},\textbf{Y}\right) = 1 - \frac{1}{N}\sum_i^N \max_j r\left(\textbf{X}_{i,j},\textbf{Y}_{i,j}\right)
\end{equation}

\noindent where $r$ is the Spearman-correlation matrix and $N$ is the number of rows in $r$.  We select a random AVIRIS HSI, generate SMCAE features from all three sensors, and then compute the dissimilarity metrics for every feature response pair (Table~\ref{table:rsa}).  Although there is some feature overlap between different SMCAE models, there is some new information that comes from combining learned features from multiple sensors.  The SMCAE models trained on Hyperion and AVIRIS are more similar than any combination with the SMCAE trained on GLiHT because AVIRIS and Hyperion span the short-wave infrared spectrum whereas GLiHT only spans through the near infrared.

\begin{table}[t]
\centering \footnotesize
\caption{Dissimilarity between the feature responses from all three SMCAE models.  The higher the value, the more dissimilar the two feature representations are.}
\begin{tabular}{@{}r|ccc@{}}\toprule
 & \textbf{AVIRIS} & \textbf{GLiHT} & \textbf{Hyperion}  \\
\midrule
\textbf{AVIRIS}   & 0.000 & 0.108 & 0.093 \\ 
\textbf{GLiHT}    &  & 0.000 & 0.105 \\ 
\textbf{Hyperion} &  & & 0.000 \\
\bottomrule 
\end{tabular}
\label{table:rsa}
\end{table}

The annotated benchmarks evaluated in this paper all have dramatically different ground sample distances (GSDs) ranging from 1.3 meters to 20 meters; yet, the state-of-the-art performance on each of these datasets could indicate that SMCAE is learning scale-invariant features.  In addition, the collection of HSI from different climates, scenes, and weather/atmosphere conditions further improve learned feature generalization; and ultimately, could enable the seamless transfer of spatial-spectral features across different sensors, environments, and machine learning tasks.

\section{Conclusion}

In this paper, we demonstrated that SMCAE learns more discriminative self-taught learning features by correcting errors in both shallow and deeper layers during training.  We have also shown that our SS-MLP classifier is effective at low-shot learning and able to handle high-dimensional inputs.  Our SuSA framework achieved state-of-the-art performance on both IEEE GRSS benchmarks for HSI semantic segmentation and have established a high bar for low-shot learning of HSI datasets.  Future work will include scaling these frameworks to other data modalities (e.g., MSI, thermal, synthetic aperture radar, etc.), higher GSD imagery (e.g., centimeter resolution imagery taken from drones), and other remote sensing tasks (e.g., target detection, crop health estimation, etc.). 

\section*{Acknowledgements}
The authors would like to thank Purdue, Pavia University, the Hysens project, and NASA/JPL-Caltech for making their remote sensing data publicly available. 

{\small
\bibliographystyle{ieee}
\bibliography{egbib}

\begin{thebibliography}{10}\itemsep=-1pt

\bibitem{bellman2013dynamic}
R.~Bellman.
\newblock {\em Dynamic programming}.
\newblock Courier Corporation, 2013.

\bibitem{spectral}
T.~Boggs.
\newblock Spectral python, 2010.

\bibitem{bruzzone2005transductive}
L.~Bruzzone, M.~Chi, and M.~Marconcini.
\newblock Transductive svms for semisupervised classification of hyperspectral
  data.
\newblock In {\em Geoscience and Remote Sensing Symposium}, volume~1, pages
  4--pp. IEEE, 2005.

\bibitem{bruzzone2006novel}
L.~Bruzzone, M.~Chi, and M.~Marconcini.
\newblock A novel transductive svm for semisupervised classification of
  remote-sensing images.
\newblock {\em Transactions on Geoscience and Remote Sensing},
  44(11):3363--3373, 2006.

\bibitem{camps2007semi}
G.~Camps-Valls, T.~V.~B. Marsheva, and D.~Zhou.
\newblock Semi-supervised graph-based hyperspectral image classification.
\newblock {\em Transactions on Geoscience and Remote Sensing},
  45(10):3044--3054, 2007.

\bibitem{clevert2015fast}
D.-A. Clevert, T.~Unterthiner, and S.~Hochreiter.
\newblock Fast and accurate deep network learning by exponential linear units
  (elus).
\newblock In {\em Proceedings of the International Conference on Learning
  Representations}, 2016.

\bibitem{dopido2014new}
I.~D{\'o}pido, J.~Li, P.~Gamba, and A.~Plaza.
\newblock A new hybrid strategy combining semisupervised classification and
  unmixing of hyperspectral data.
\newblock {\em Journal of Selected Topics in Applied Earth Observations and
  Remote Sensing}, 7(8):3619--3629, 2014.

\bibitem{dozat2016incorporating}
T.~Dozat.
\newblock Incorporating nesterov momentum into adam.
\newblock 2016.

\bibitem{ghamisi2014automatic}
P.~Ghamisi et~al.
\newblock Automatic framework for spectral--spatial classification based on
  supervised feature extraction and morphological attribute profiles.
\newblock {\em Journal of Selected Topics in Applied Earth Observations and
  Remote Sensing}, 7(6):2147--2160, 2014.

\bibitem{glorot2010understanding}
X.~Glorot and Y.~Bengio.
\newblock Understanding the difficulty of training deep feedforward neural
  networks.
\newblock In {\em Proceedings of the International Conference on Artificial
  Intelligence and Statistics}, pages 249--256, 2010.

\bibitem{gomez2008semisupervised}
L.~G{\'o}mez-Chova, G.~Camps-Valls, J.~Munoz-Mari, and J.~Calpe.
\newblock Semisupervised image classification with laplacian support vector
  machines.
\newblock {\em Geoscience and Remote Sensing Letters}, 5(3):336--340, 2008.

\bibitem{imani2013boundary}
M.~Imani and H.~Ghassemian.
\newblock Boundary based supervised classification of hyperspectral images with
  limited training samples.
\newblock {\em ISPRS-International Archives of the Photogrammetry, Remote
  Sensing and Spatial Information Sciences}, (3):203--207, 2013.

\bibitem{kemker2017self}
R.~Kemker and C.~Kanan.
\newblock Self-taught feature learning for hyperspectral image classification.
\newblock {\em Transactions on Geoscience and Remote Sensing},
  55(5):2693--2705, 2017.

\bibitem{kriegeskorte2008representational}
N.~Kriegeskorte, M.~Mur, and P.~Bandettini.
\newblock Representational similarity analysis--connecting the branches of
  systems neuroscience.
\newblock {\em Frontiers in systems neuroscience}, 2, 2008.

\bibitem{lin2013spectral}
Z.~Lin, Y.~Chen, X.~Zhao, and G.~Wang.
\newblock Spectral-spatial classification of hyperspectral image using
  autoencoders.
\newblock In {\em Information, Communications and Signal Processing}, pages
  1--5. IEEE, 2013.

\bibitem{liu2017semi}
B.~Liu, X.~Yu, P.~Zhang, X.~Tan, A.~Yu, and Z.~Xue.
\newblock A semi-supervised convolutional neural network for hyperspectral
  image classification.
\newblock {\em Remote Sensing Letters}, 8(9):839--848, 2017.

\bibitem{liu2015hyperspectral}
Y.~Liu, G.~Cao, Q.~Sun, and M.~Siegel.
\newblock Hyperspectral classification via deep networks and superpixel
  segmentation.
\newblock {\em International Journal of Remote Sensing}, 36(13):3459--3482,
  2015.

\bibitem{ma2015hyperspectral}
X.~Ma, J.~Geng, and H.~Wang.
\newblock Hyperspectral image classification via contextual deep learning.
\newblock {\em EURASIP Journal on Image and Video Processing}, 2015(1):20,
  2015.

\bibitem{maas2013rectifier}
A.~L. Maas, A.~Y. Hannun, and A.~Y. Ng.
\newblock Rectifier nonlinearities improve neural network acoustic models.
\newblock In {\em Proceedings of the International Conference on Machine
  Learning}, volume~30, 2013.

\bibitem{mirzapour2015improving}
F.~Mirzapour and H.~Ghassemian.
\newblock Improving hyperspectral image classification by combining spectral,
  texture, and shape features.
\newblock {\em International Journal of Remote Sensing}, 36(4):1070--1096,
  2015.

\bibitem{raina2007self}
R.~Raina, A.~Battle, H.~Lee, B.~Packer, and A.~Y. Ng.
\newblock Self-taught learning: transfer learning from unlabeled data.
\newblock In {\em Proceedings of the International Conference on Machine
  Learning}, pages 759--766. ACM, 2007.

\bibitem{rasmus2015semi}
A.~Rasmus, H.~Valpola, M.~Honkala, M.~Berglund, and T.~Raiko.
\newblock Semi-supervised learning with ladder networks.
\newblock In {\em Advances in Neural Information Processing Systems}, pages
  3546--3554, 2015.

\bibitem{ratle2010semisupervised}
F.~Ratle, G.~Camps-Valls, and J.~Weston.
\newblock Semisupervised neural networks for efficient hyperspectral image
  classification.
\newblock {\em Transactions on Geoscience and Remote Sensing},
  48(5):2271--2282, 2010.

\bibitem{salimans2016improved}
T.~Salimans, I.~Goodfellow, W.~Zaremba, V.~Cheung, A.~Radford, and X.~Chen.
\newblock Improved techniques for training gans.
\newblock In {\em Advances in Neural Information Processing Systems}, pages
  2234--2242, 2016.

\bibitem{shen2013discriminative}
L.~Shen et~al.
\newblock Discriminative gabor feature selection for hyperspectral image
  classification.
\newblock {\em Geoscience and Remote Sensing Letters}, 10(1):29--33, 2013.

\bibitem{soltani2015pixels}
A.~Soltani-Farani and H.~R. Rabiee.
\newblock When pixels team up: spatially weighted sparse coding for
  hyperspectral image classification.
\newblock {\em Geoscience and Remote Sensing Letters}, 12(1):107--111, 2015.

\bibitem{tang2015hyperspectral}
Y.~Y. Tang, Y.~Lu, and H.~Yuan.
\newblock Hyperspectral image classification based on three-dimensional
  scattering wavelet transform.
\newblock {\em Transactions on Geoscience and Remote Sensing},
  53(5):2467--2480, 2015.

\bibitem{tao2015unsupervised}
C.~Tao, H.~Pan, Y.~Li, and Z.~Zou.
\newblock Unsupervised spectral--spatial feature learning with stacked sparse
  autoencoder for hyperspectral imagery classification.
\newblock {\em Geoscience and Remote Sensing Letters}, 12(12):2438--2442, 2015.

\bibitem{trottier-pelu}
L.~Trottier, P.~Gigu{\`{e}}re, and B.~Chaib{-}draa.
\newblock Parametric exponential linear unit for deep convolutional neural
  networks.
\newblock {\em CoRR}, abs/1605.09332, 2016.

\bibitem{valpola2015neural}
H.~Valpola.
\newblock From neural {PCA} to deep unsupervised learning.
\newblock {\em Advances in Independent Component Analysis and Learning
  Machines}, pages 143--171, 2015.

\bibitem{yang2017sparse}
L.~Yang, M.~Wang, S.~Yang, R.~Zhang, and P.~Zhang.
\newblock Sparse spatio-spectral lapsvm with semisupervised kernel propagation
  for hyperspectral image classification.
\newblock {\em Journal of Selected Topics in Applied Earth Observations and
  Remote Sensing}, 10(5):2046--2054, 2017.

\bibitem{yang2014semi}
L.~Yang, S.~Yang, P.~Jin, and R.~Zhang.
\newblock Semi-supervised hyperspectral image classification using
  spatio-spectral laplacian support vector machiney.
\newblock {\em Geoscience and Remote Sensing Letters}, 11(3):651--655, 2014.

\bibitem{zhao2015combining}
W.~Zhao, Z.~Guo, J.~Yue, X.~Zhang, and L.~Luo.
\newblock On combining multiscale deep learning features for the classification
  of hyperspectral remote sensing imagery.
\newblock {\em International Journal of Remote Sensing}, 36(13):3368--3379,
  2015.

\end{thebibliography}
}

\begin{IEEEbiography}[{\includegraphics[width=1in,height=1.25in,clip,keepaspectratio]{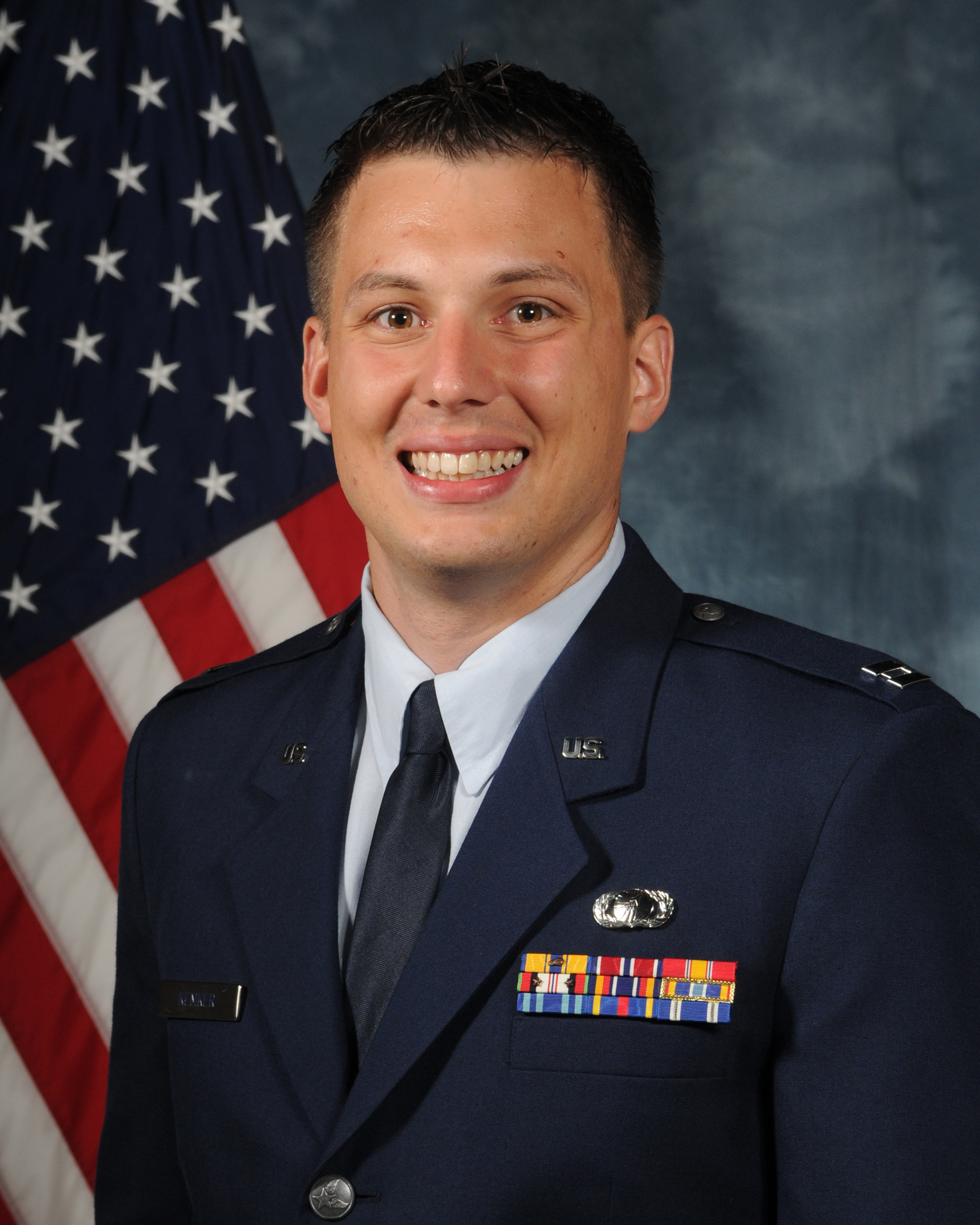}}]{Ronald Kemker} is a PhD Candidate in the Chester F. Carlson Center for Imaging Science at the Rochester Institute of Technology.  His research currently involves applying computer vision and machine learning techniques to tackle remote sensing problems.  He received his MS degree in Electrical Engineering from Michigan Technological University.
\end{IEEEbiography}

\begin{IEEEbiographynophoto}{Ryan Luu} is a student at Victor Senior High School and an intern at the Chester F. Carlson Center for Imaging Science at the Rochester Institute of Technology. His current projects involve object detection and robot navigation.
\end{IEEEbiographynophoto}

\begin{IEEEbiography}[{\includegraphics[width=1in,height=1.25in,clip,keepaspectratio]{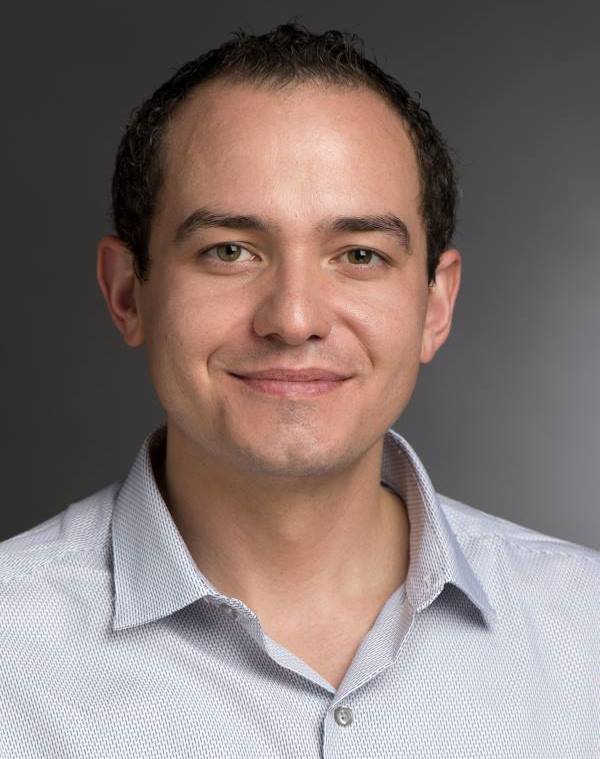}}]{Christopher Kanan} is an assistant professor in the Chester F. Carlson Center for Imaging Science at the Rochester Institute of Technology. His lab applies deep learning to problems in computer vision. His recent projects have focused on object recognition, object detection, active vision, visual question answering,  semantic segmentation, and lifelong machine learning. He received a PhD in computer science from the University of California at San Diego and a MS in computer science from the University of Southern California.
\end{IEEEbiography}

\end{document}